\documentclass[letterpaper, 10 pt, journal, twoside]{IEEEtran}
\IEEEoverridecommandlockouts                              

\usepackage[utf8]{inputenc}
\usepackage[T1]{fontenc}
\usepackage{times}
\usepackage{amsfonts}
\usepackage{amsmath}
\usepackage{amssymb}

\usepackage{amsthm}
\usepackage{mathtools}
\usepackage{bm}

\usepackage[english]{babel}
\addto\captionsenglish{\renewcommand{\figurename}{Fig.}}
\usepackage{graphicx}
\usepackage{svg}
\usepackage{epsfig}
\usepackage{psfrag}

\usepackage{array}
\usepackage{multirow}
\usepackage{booktabs}
\usepackage{makecell}
\usepackage{tabularx}
\usepackage[flushleft]{threeparttable}

\usepackage{float}
\usepackage{placeins}
\usepackage{stfloats}
\usepackage{cuted}
\usepackage{multicol}
\usepackage[figuresright]{rotating}
\usepackage[caption=false,font=small,labelfont=sf,textfont=sf]{subfig}
\usepackage[font=small,labelfont=bf]{caption}
\usepackage{algorithm}
\usepackage{algpseudocode}

\usepackage{xcolor}
\usepackage{xspace}
\usepackage{siunitx}
\usepackage{fancyhdr}
\usepackage{pifont}
\usepackage{textcomp}
\usepackage{cancel}
\usepackage{dirtytalk}
\usepackage{comment}
\usepackage{verbatim}
\usepackage{xparse}
\usepackage{listings}
\usepackage{calc}
\usepackage[nolist]{acronym}
\usepackage{cite}

\usepackage[hidelinks]{hyperref}
\usepackage{url}
\usepackage{macros}

\makeatletter
\def\fnum@table{Table~\thetable}
\long\def\@makecaption#1#2{%
\ifx\@captype\@IEEEtablestring%
\setbox\@tempboxa\hbox{\footnotesize #1.~~ #2}%
\ifdim \wd\@tempboxa >\hsize%
\setbox\@tempboxa\hbox{\footnotesize #1.~~ }%
\parbox[t]{\hsize}{\footnotesize \noindent\unhbox\@tempboxa#2}%
\else%
\hbox to\hsize{\footnotesize\box\@tempboxa\hfil}%
\fi%
\@IEEEtablecaptionsepspace%
\else%
\@IEEEfigurecaptionsepspace%
\setbox\@tempboxa\hbox{\footnotesize #1.~~ #2}%
\ifdim \wd\@tempboxa >\hsize%
\setbox\@tempboxa\hbox{\footnotesize #1.~~ }%
\parbox[t]{\hsize}{\footnotesize \noindent\unhbox\@tempboxa#2}%
\else%
\ifCLASSOPTIONconference \hbox to\hsize{\footnotesize\hfil\box\@tempboxa\hfil}%
\else \hbox to\hsize{\footnotesize\box\@tempboxa\hfil}%
\fi\fi\fi}
\makeatother

\newcommand{\best}[1]{\textbf{#1}}
\newcommand{\second}[1]{\underline{#1}}

\newcommand{\figref}[1]{\hyperref[#1]{Fig.~\ref*{#1}}}
\newcommand{\tabref}[1]{\hyperref[#1]{Table~\ref*{#1}}}

\fancypagestyle{ralpages}{%
  \fancyhf{}%
  \fancyhead[L]{\ifodd\value{page}\else\scriptsize\thepage\fi}%
  \fancyhead[R]{\ifodd\value{page}\scriptsize\thepage\fi}%
}
\DeclareUnicodeCharacter{2013}{--}
\DeclareUnicodeCharacter{2014}{---}
\DeclareUnicodeCharacter{2019}{'}

\newcommand{\scenariob}[1]{{\fontsize{15}{15}\selectfont\sf#1}\xspace}
\newcommand{\scenariot}[1]{{\fontsize{8}{8}\selectfont\sf#1}\xspace}

\newcommand{\scenariof}[1]{{\fontsize{7}{7}\selectfont\sf#1}\xspace}
\newcommand{\gofusion}{\scenariot{GOFusion}}
\newcommand{\gofusioncaption}{\scenariof{GOFusion}}
\newcommand{\okvis}{OKVIS2-X\xspace}

\title{\LARGE \bf
\scenariob{GOFusion:} Real-Time Gaussian-Occupancy Submap Fusion for \\ Fisher-Informed Autonomous Robot Navigation
}

\author{Boxin Bu$^{1,2}$, Simon Boche$^{1}$, Haoming Zhang$^{2}$, Jaehyung Jung$^{1}$, Jiaxin Wei$^{1}$, Hannah Schieber$^{3}$, \\Daniel Roth$^{3}$, Angela P. Schoellig$^{2,4}$, and Stefan Leutenegger$^{1}$
\thanks{$^{1}$Mobile Robotics Lab, Technical University of Munich and ETH Zurich.}
\thanks{$^{2}$Learning Systems \& Robotics Lab, Technical University of Munich and Institute of Robotics and Machine Intelligence (MIRMI).}%
\thanks{$^{3}$Human-Centered Computing and Extended Reality Lab,
TUM University Hospital, Clinic for Orthopedics and Sports Orthopedics, 
Munich Institute of Robotics and Machine Intelligence (MIRMI).}
\thanks{$^{4}$University of Toronto Institute for Aerospace Studies, University of Toronto Robotics Institute, and Vector Institute for Artificial Intelligence, and Robotics Institute Germany.}
\thanks{E-mail addresses: \{first.secondname\}@tum.de}
}

\fancypagestyle{arxivnotice}{%
  \fancyhf{}%
  \fancyhead[C]{\normalfont\scriptsize This work has been submitted to the IEEE for possible publication. Copyright may be transferred without notice, after which this version may no longer be accessible.}%
  \fancyhead[R]{\normalfont\scriptsize\thepage}%
}

\begin{document}
\bstctlcite{IEEEexample:BSTcontrol}
\begin{acronym}[Bspwwww.]  

\acro{ar}[AR]{augmented reality}
\acro{xr}[XR]{extended reality}
\acro{ap}[AP]{average precision}
\acro{api}[API]{application programming interface}
\acroplural{ann}[ANN]{artifical neural networks}
\acro{bev}[BEV]{bird eye view}
\acro{rbob}[BRB]{Bottleneck residual block}
\acroplural{rbob}[BRBs]{Bottleneck residual blocks}
\acro{mbiou}[mBIoU]{mean Boundary Intersection over Union}
\acro{cai}[CAI]{computer-assisted intervention}
\acro{ce}[CE]{cross entropy}
\acro{cad}[CAD]{computer-aided design}
\acro{cnn}[CNN]{convolutional neural network}

\acro{crf}[CRF]{conditional random fields}
\acro{dpc}[DPC]{dense prediction cells}
\acro{dla}[DLA]{deep layer aggregation}
\acro{dnn}[DNN]{deep neural network}
\acroplural{dnn}[DNNs]{deep neural networks}

\acro{da}[DA]{domain adaption}
\acro{dr}[DR]{domain randomization}
\acro{fat}[FAT]{falling things}
\acro{fps}[FPS]{frames per second}
\acro{fcn}[FCN]{fully convolutional network}
\acroplural{fcn}[FCNs]{fully convolutional networks}
\acro{fov}[FoV]{field of view}
\acro{fv}[FV]{front view}
\acro{fp}[FP]{False Positive}
\acro{fpn}[FPN]{feature Pyramid network}
\acro{fn}[FN]{False Negative}
\acro{fmss}[FMSS]{fast motion sickness scale}
\acro{gan}[GAN]{generative adversarial network}
\acroplural{gan}[GANs]{generative adversarial networks}
\acro{gcn}[GCN]{graph convolutional network}
\acroplural{gcn}[GCNs]{graph convolutional networks}
\acro{gs}[GS]{Gaussian Splatting}
\acro{3dgs}[3DGS]{3D Gaussian Splatting}
\acro{gg}[GG]{Gaussian Grouping}
\acro{hmi}[HMI]{Human-Machine-Interaction}
\acro{hmd}[HMD]{Head Mounted Display}
\acroplural{hmd}[HMDs]{head mounted displays}
\acro{iou}[IoU]{intersection over union}
\acro{irb}[IRB]{inverted residual bock}
\acroplural{irb}[IRBs]{inverted residual blocks}
\acro{ipq}[IPQ]{igroup presence questionnaire}
\acro{knn}[KNN]{k-nearest-neighbor}
\acro{lidar}[LiDAR]{light detection and ranging}
\acro{lsfe}[LSFE]{large scale feature extractor}
\acro{llm}[LLM]{large language model}
\acro{map}[mAP]{mean average precision}
\acro{mc}[MC]{mismatch correction module}
\acro{miou}[mIoU]{mean intersection over union}
\acro{mis}[MIS]{Minimally Invasive Surgery}
\acro{msdl}[MSDL]{Multi-Scale Dice Loss}
\acro{ml}[ML]{Machine Learning}
\acro{mlp}[MLP]{multilayer perception}
\acro{miou}[mIoU]{mean Intersection over Union}
\acro{nn}[NN]{neural network}
\acroplural{nn}[NNs]{neural networks}
\acro{ndd}[NDDS]{NVIDIA Deep Learning Data Synthesizer}
\acro{nocs}[NOCS]{Normalized Object Coordiante Space}
\acro{nerf}[NeRF]{Neural Radiance Fields}
\acro{NVISII}[NVISII]{NVIDIA Scene Imaging Interface}
\acro{ngp}[NGP]{Neural Graphics Primitives}
\acro{or}[OR]{Operating Room}
\acro{pbr}[PBR]{physically based rendering}
\acro{psnr}[PSNR]{peak signal-to-noise ratio}
\acro{pnp}[PnP]{Perspective-n-Point}
\acro{rv}[RV]{range view}
\acro{roi}[ROI]{region of interest}
\acroplural{roi}[ROIs]{region of interests}
\acro{rbab}[BB]{residual basic block}
\acro{ras}[RAS]{robot-assisted surgery}
\acroplural{rbab}[BBs]{residual basic blocks}
\acro{spp}[SPP]{spatial pyramid pooling}
\acro{sh}[SH]{spherical harmonics}
\acro{sgd}[SGD]{stochastic gradient descent}
\acro{sdf}[SDF]{signed distance field}
\acro{sfm}[SfM]{structure-from-motion}
\acro{sam}[SAM]{Segment-Anything}
\acro{sus}[SUS]{system usability scale}
\acro{ssim}[SSIM]{structural similarity index measure}
\acro{sfm}[SfM]{structure from motion}
\acro{slam}[SLAM]{simultaneous localization and mapping}
\acro{tp}[TP]{True Positive}
\acro{tn}[TN]{True Negative}
\acro{thor}[thor]{The House Of inteRactions}
\acro{tsdf}[TSDF]{truncated signed distance function}
\acro{vr}[VR]{Virtual Reality}
\acro{ycb}[YCB]{Yale-CMU-Berkeley}

\acro{ar}[AR]{augmented reality}
\acro{ate}[ATE]{absolute trajectory error}
\acro{bvip}[BVIP]{blind or visually impaired people}
\acro{cnn}[CNN]{convolutional neural network}
\acro{c2f}[c2f]{coarse-to-fine}
\acro{fov}[FoV]{field of view}
\acro{gan}[GAN]{generative adversarial network}
\acro{gcn}[GCN]{graph convolutional Network}
\acro{gnn}[GNN]{Graph Neural Network}
\acro{hmi}[HMI]{Human-Machine-Interaction}
\acro{hmd}[HMD]{head-mounted display}
\acro{mr}[MR]{mixed reality}
\acro{iot}[IoT]{internet of things}
\acro{llff}[LLFF]{Local Light Field Fusion}
\acro{bleff}[BLEFF]{Blender Forward Facing}

\acro{lpips}[LPIPS]{learned perceptual image patch similarity}
\acro{nerf}[NeRF]{neural radiance fields}
\acro{nvs}[NVS]{novel view synthesis}
\acro{nbv}[NBV]{next best view}
\acro{mlp}[MLP]{multilayer perceptron}
\acro{mrs}[MRS]{Mixed Region Sampling}

\acro{or}[OR]{Operating Room}
\acro{pbr}[PBR]{physically based rendering}
\acro{psnr}[PSNR]{peak signal-to-noise ratio}
\acro{pnp}[PnP]{Perspective-n-Point}
\acro{pc}[PC]{point cloud}
%
\acro{sus}[SUS]{system usability scale}
\acro{ssim}[SSIM]{structural similarity index measure}
\acro{sfm}[SfM]{structure from motion}
\acro{slam}[SLAM]{simultaneous localization and mapping}

\acro{tp}[TP]{True Positive}
\acro{tn}[TN]{True Negative}
\acro{thor}[thor]{The House Of inteRactions}
\acro{ueq}[UEQ]{User Experience Questionnaire}
\acro{vr}[VR]{virtual reality}
\acro{who}[WHO]{World Health Organization}
\acro{xr}[XR]{extended reality}
\acro{ycb}[YCB]{Yale-CMU-Berkeley}
\acro{yolo}[YOLO]{you only look once}
\acro{vb}[VB]{voting-based}
\end{acronym}
\maketitle
\thispagestyle{arxivnotice}

\begin{abstract}


In this letter, we present \gofusion, a real-time visual mapping system for autonomous robot navigation that anchors Gaussian primitives to local occupancy maps, providing both spatial awareness for path planning and high-fidelity appearance that enables information-driven robot navigation. \gofusion builds 3D Gaussian maps from pose estimates and local occupancy maps obtained via visual-inertial \ac{slam} and enables pose corrections to maintain both geometric and visual consistency. Furthermore, it restricts the 3D Gaussian Splatting optimization to active submaps to bound GPU memory usage. Building on the appearance-rich Gaussian map, we introduce a rasterizer-integrated Fisher approximation that quantifies per-primitive sensitivity online, serving as an information-driven heuristic for Gaussian pruning and next-best-view planning. We evaluate \gofusion on public datasets and in real-world drone flights. It not only outperforms several baselines in mapping quality, but also achieves real-time, high-fidelity and uncertainty-aware mapping at scale. Finally, we demonstrate fully autonomous drone exploration for Gaussian map refinement using the proposed Fisher-informed heuristic. Code is available at \url{https://github.com/Bennnoo/GOFusion}. We provide a supplementary video demonstration at \url{https://youtu.be/pLQKbxQvb6I}.

\end{abstract}

\section{Introduction}\label{sec:introduction}

Autonomous robots need maps that support safe motion, reliable localization, and visual understanding in unknown environments. 
Occupancy maps~\cite{moravec1985high,elfes1989occupancy,hornung2013octomap} distinguish between free, occupied, and unknown space so that planners can check collisions, extract frontiers~\cite{yamauchi1997frontier}, and generate paths in large environments~\cite{papatheodorou2025efficient}. 
Such maps, however, encode geometry alone and discard appearance, whereas advanced tasks such as inspection and human-robot interaction require detailed scene appearance. 

Radiance field map representations address this limitation, either by implicitly encoding the scene in a neural network~\cite{mildenhall2020nerf} or by explicitly optimizing Gaussian primitives via \ac{3dgs}~\cite{kerbl2023gaussian}. While the former relies on dense per-ray sampling of an implicit field and therefore remains costly to train and query, \ac{3dgs} supports efficient rendering via rasterization of an explicit set of primitives and is therefore better suited to online mapping. However, vanilla \ac{3dgs} maps describe only observed surfaces and do not distinguish free space from potentially occupied space, which is unsafe for robot navigation.

\begin{figure}[!t]
    \centering
    \includegraphics[width=1\linewidth]{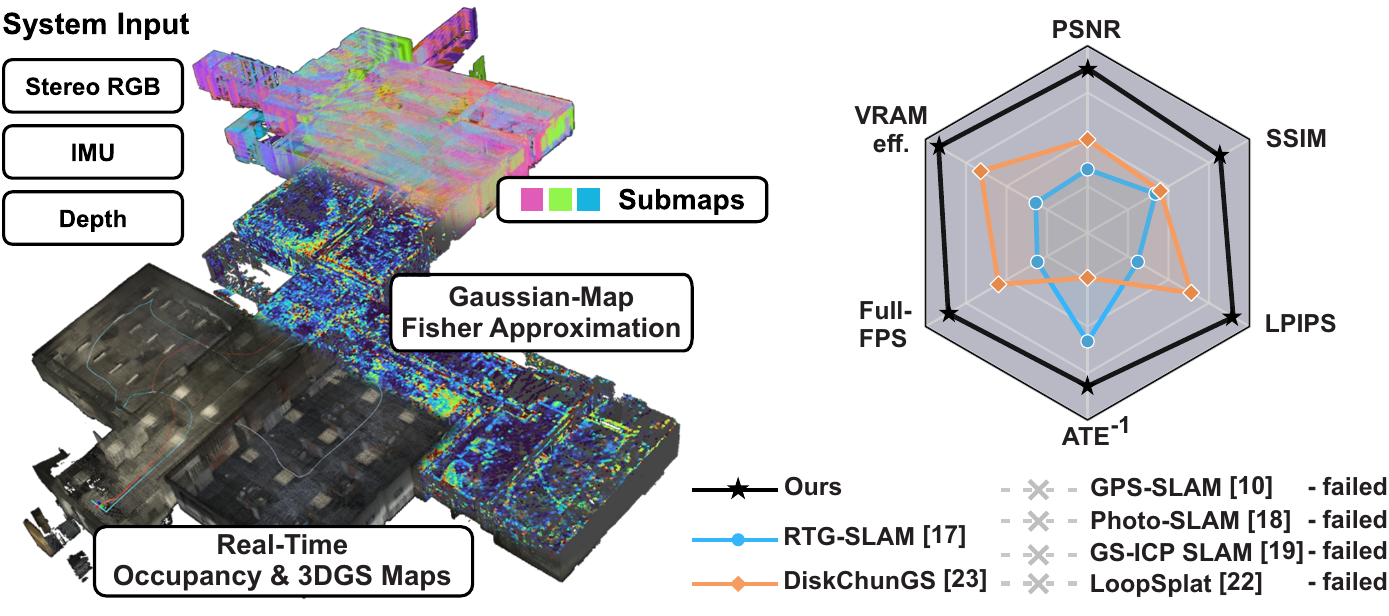}
    \caption{Benchmark result on the uHumans2 dataset~\cite{rosinol2021kimera}.
    Left: finalized map of \gofusioncaption, presenting different submaps. 
    Right: performance summary of \gofusioncaption 
    on rendering quality (PSNR/SSIM/LPIPS), trajectory accuracy ($\mathrm{ATE}^{-1}$), speed (Full-FPS), and GPU-memory efficiency (VRAM eff.).}
    \label{fig:uhumans2_demo}
\end{figure}

In principle, combining occupancy and Gaussian maps yields a complementary hybrid representation, in which an occupancy layer supports free-space reasoning, completeness assessment, and planning, while a Gaussian layer provides high-fidelity appearance. However, existing systems realize only half of this potential. GSFusion~\cite{wei2024gsfusion} and GPS-SLAM~\cite{peng2025gpsslam} exploit this coupling in one direction, predominantly using the geometric map to improve the reconstruction quality and efficiency rather than to support robot localization and navigation. Therefore, a hybrid representation that directly supports navigation remains an open challenge.

To address this challenge, we propose \gofusion, a real-time hybrid Gaussian-and-occupancy mapping system for autonomous robot navigation. \gofusion is built directly on a real-time visual-inertial \ac{slam} system, \okvis~\cite{boche2025okvis2x}, and inherits its state estimates and occupancy submaps. However, unlike previous reconstruction-oriented methods, we close the loop between the two map layers. While the occupancy submaps anchor and initialize the Gaussians, photometric optimization of \ac{3dgs}, in turn, refines the camera poses, thereby reducing pose jitter and improving map consistency. Our submap-local design keeps online optimization scalable, since trainable states are maintained only for active submaps. Beyond geometry and appearance, we further accumulate Fisher information from the Gaussian primitives to identify poorly-constrained map regions, 
providing an information-theoretic heuristic for trajectory planning that enables the robot to actively complete its own map. Fig.\,\ref{fig:uhumans2_demo} presents the performance of our method compared to various baselines. 

To our knowledge, \gofusion is the first hybrid Gaussian-occupancy mapping system that supports real-time 3D navigation outdoors from cameras and an IMU alone. The contributions of this paper are:

\begin{enumerate}
    \item A submap-based 3D Gaussian representation for large-scale high-fidelity mapping that supports appearance-dependent tasks such as inspection and teleoperation alongside navigation, where \ac{slam} pose-graph updates, including loop closures, also directly update both the occupancy and Gaussian layers.

    \item A rasterizer-integrated online Fisher approximation that maintains a diagonal, accumulated Fisher information over all Gaussians at negligible cost. Serving as an information heuristic for robot exploration, it highlights another key methodological advantage of Gaussian mapping in robotics.
    
    \item A real-time navigation framework that integrates visual-inertial SLAM with our hybrid map, deployable either fully onboard or split across onboard and offboard compute for resource-limited platforms.

    \item Real-world UAV experiments demonstrate \gofusion's online capabilities and how Fisher information can be used in Gaussian pruning and next-best-view selection. 
\end{enumerate}

\section{Related Work}\label{sec:related_work}

\subsection{Volumetric Mapping for Robot Navigation}
Volumetric representations have dominated robot mapping for decades. 
Occupancy maps~\cite{moravec1985high} 
explicitly distinguish free, occupied, and unknown space, supporting collision checking, frontier extraction~\cite{yamauchi1997frontier}, and navigation under partial observation~\cite{ryll_partially_planning}. \Ac{tsdf}- and \ac{sdf}-based systems such as \cite{newcombe2011kinectfusion,oleynikova2017voxblox} instead target continuous surface reconstruction. They encode distance to the nearest observed surface, so unobserved volume is not separated from free space by the representation itself and must be handled by convention. Octree frameworks such as \cite{hornung2013octomap} improve scalability through hierarchical subdivision. These methods scale in space but not in time, since a globally parameterized map must be re-optimized whenever loop closure revises past poses.

Submap-based SLAM resolves this issue by integrating measurements into local maps, whose poses are frequently updated during graph optimization, 
improving global map consistency across all maps. This concept lends itself to exploration in large environments, where local submaps enable efficient updates while still supporting global planning~\cite{papatheodorou2025efficient}.
\okvis~\cite{boche2025okvis2x} follows this direction for multi-sensor state estimation with occupancy submapping and submap alignment. Because pose corrections act at the submap level, the occupancy map remains directly available for planning throughout. 

These advantages, however, concern geometry alone. As occupancy quantifies uncertainty about space, not appearance, a surface can be reliable for collision avoidance yet unusable for inspection or human-robot collaboration.

\subsection{Gaussian Mapping for Robot Navigation}
A more recent line of work in mapping replaces volumetric structures with explicit, addable, prunable, and rigidly transformable Gaussian primitives. 
Moreover, they are also differentiable under a fast rasterizer. Supervision is derived directly from image intensities, so every pixel contributes a residual~\cite{kerbl2023gaussian}. 
Dense visual SLAM systems exploit this to jointly optimize camera poses and Gaussian primitives~\cite{keetha2024splatam, matsuki2024gaussian}, or to embed Gaussians in online pipelines through selective primitive updates, conventional frontends, and RGB-D alignment~\cite{peng2024rtgslam, huang2024photoslam, ha2024rgbdgsicp}. Others pair Gaussian maps with dedicated planners or semantic objectives for closed-loop navigation~\cite{chen2025splatnav, nagami2026vista}. Hybrid systems go one step further: previous work \cite{wei2024gsfusion, peng2025gpsslam} use \ac{tsdf} geometry to guide Gaussian insertion and suppress redundant primitives, although the geometry serves reconstruction rather than a trajectory planner.

At the robot scale, Gaussian SLAM faces two further obstacles. A single global Gaussian map carries optimizer states, gradients, and rendering buffers that grow with scene size, and, once built, cannot easily adopt pose corrections. 
LoopSplat~\cite{zhu2025loopsplat} addresses the second by organizing Gaussians into submaps that loop closure can rigidly transform, while DiskChunGS~\cite{feldmann2026diskchungs} addresses the first with disk-backed chunking. Neither, however, maintains an occupancy representation for downstream robotic tasks. 

\subsection{Uncertainty-Aware Mapping for Robot Navigation}
Mapping to support real-world robot navigation, however, requires more than geometry and appearance, since the robot must also decide where to go next. Frontier-based exploration~\cite{yamauchi1997frontier, papatheodorou2025efficient} exposes unknown regions, but is blind to observed regions where the appearance model remains poorly constrained. Neural and Gaussian methods therefore study uncertainty-aware view selection, using Fisher information~\cite{jiang2024fisherrf} or related visibility and information-gain signals~\cite{jin2025activegs, chen2025activegamer} to improve reconstruction quality. A second line keeps uncertainty inside the estimator, reweighting tracking, registration, and loop constraints~\cite{tran2026varsplat}. 

Closest to our setting, $\mu$SLAM~\cite{jiang2026muslam} maintains a per-primitive confidence and plans on an occupancy grid projected from its Gaussian field. Its uncertainty never enters the planner, and the projected grid supports no planning beyond the ground plane. 
None of these systems, however, couples map uncertainty with the free-space reasoning required for real-time robot exploration. This motivates \gofusion's unified hybrid map representation that keeps Gaussian optimization local to the active submap while exposing a per-primitive uncertainty for downstream decisions. 

\section{Method}\label{sec:method}

\begin{figure*}[t]
    \centering
    \includegraphics[width=\textwidth]{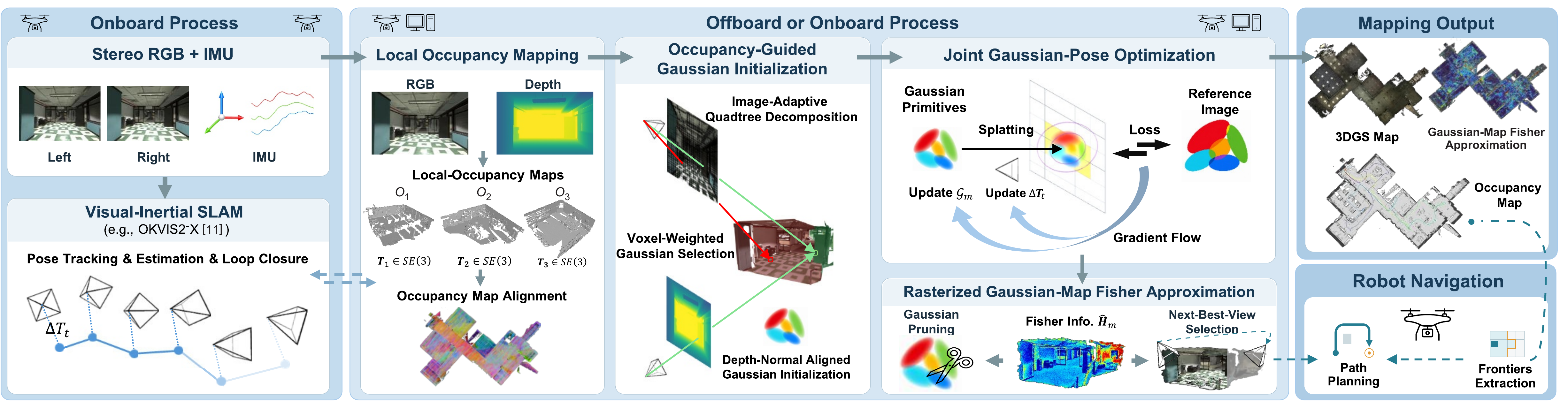}
    \caption{System overview of \gofusioncaption. OKVIS2-X~\cite{boche2025okvis2x} processes stereo RGB, IMU, and depth to estimate poses and build local-occupancy maps. Each occupancy map guides image-adaptive, voxel-weighted, and normal-aligned Gaussian initialization. Online Gaussian optimization then refines the active submap and provides gradient buffers to compute the accumulated Fisher approximation, which can be used for redundant Gaussian pruning or next-best-view selection. Our system can run entirely onboard or split between onboard and offboard processes, depending on the available compute.}
    \label{fig:method_overview}
    \vspace{-0.3cm}
\end{figure*}

Throughout this paper, we denote the world frame and the local frame of a local-occupancy map $\myFrameVec{O}{}{m}{}$ by $\underrightarrow{\mathcal{F}}_W$ and $\underrightarrow{\mathcal{F}}_{O_m}$, respectively, and a transformation $\myFrameVec{T}{}{AB}{}\in SE(3)$ represents the rigid transform from $\underrightarrow{\mathcal{F}}_B$ to $\underrightarrow{\mathcal{F}}_A$. $\myFrameVec{p}{A}{}{}$ denotes a point $\myFrameVec{p}{}{}{}$ in its reference frame $\underrightarrow{\mathcal{F}}_A$.

Targeting real-world robot navigation, we build \gofusion on top of the real-time visual-inertial SLAM system, \okvis~\cite{boche2025okvis2x}. Our mapping system takes as input its $M$ local-occupancy maps $\mathcal{O}=\{\myFrameVec{O}{}{m}{}\}^M_{m=1}$ with their corresponding pose estimates $\myFrameVec{T}{}{W O_m}{}$, together with rectified RGB and depth images, as illustrated in Fig.\,\ref{fig:method_overview}. Here, $\underrightarrow{\mathcal{F}}_{O_m}$ coincides with the pose of the anchoring keyframe of the local-occupancy map. Loop-closure correction and local map alignment run in parallel, keeping the local-occupancy maps consistent at runtime.


\subsection{Hybrid Occupancy-Gaussian Mapping}\label{subsec:hybrid_submaps}
In \gofusion, we define the complete map as a collection of hybrid submaps as:
\begin{equation}
\small
\mathcal M=\{\mathcal M_m\}_{m=1}^{M},~
\mathcal M_m=
\left(
\myFrameVec{T}{}{WS_m}{},~
\myFrameVec{O}{}{m}{},~
\myFrameVec{G}{}{m}{},~
\myFrameVecHat{H}{}{m}{}
\right),
\label{eq:GOFusion_submap_state}
\end{equation}
where we introduce the submap frame $\underrightarrow{\mathcal{F}}_{S_m} \equiv \underrightarrow{\mathcal{F}}_{O_m}$, the origin for both, the local-occupancy map $\myFrameVec{O}{}{m}{}$ as well as corresponding Gaussian primitives $\myFrameVec{G}{}{m}{}$. 
Additionally, the matrix $\myFrameVecHat{H}{}{m}{}$ stores their accumulated Fisher approximation (see Sec.\,\ref{subsec:fisher_cache_method}).

We parameterize the $i$-th Gaussian primitive $\myFrameVec{g}{}{i}{}\in\myFrameVec{G}{}{m}{}$ as
\begin{equation}
\myFrameVec{g}{}{i}{} =
\left(
\myFrameVec{\mu}{}{i}{},~
\myFrameVec{q}{}{i}{},~
\myFrameVec{s}{}{i}{},~
\myFrameScalar{\omega}{i}{},~
\myFrameVec{f}{}{i}{\rm dc},~
\myFrameVec{f}{}{i}{\rm rest}
\right)\in\mathbb R^{59},
\label{eq:gaussian_param_state}
\end{equation}
where $\myFrameVec{\mu}{}{i}{}\in\mathbb R^{3}$ and $\myFrameVec{q}{}{i}{}\in\mathbb R^{4}$ denote the center and the rotation quaternion of the Gaussian primitive in the local frame $\underrightarrow{\mathcal{F}}_{S_m}$. $\myFrameVec{s}{}{i}{}\in\mathbb R^{3}$ and $\omega_i$ represent the log-scales and the opacity logit, while $\myFrameVec{f}{}{i}{\rm dc}\in\mathbb R^3$ and
$\myFrameVec{f}{}{i}{\rm rest}\in\mathbb R^{45}$ are the
degree-3 spherical-harmonic color coefficients as in~\cite{kerbl2023gaussian}.

Our submap-local parameterization applies loop-closure correction or 3DGS pose refinement (see Sec.\,\ref{subsec:pose_feedback}) directly to the submap poses $\myFrameVec{T}{}{WS_m}{}$, leaving the Gaussian parameters untouched. These pose updates propagate to the local-occupancy map and its Gaussian primitives without re-optimization. For instance, the center $\myFrameVec{\mu}{\myFrameVec{S}{}{m}{}}{i}{}$ of the $i$-th Gaussian primitive maps to the world frame as $\myFrameVec{\mu}{W}{i}{} = \myFrameVec{T}{}{WS_m}{}\myFrameVec{\mu}{S_m}{i}{}$. Thus, 3DGS optimization remains local to the active submap, keeping only one submap in a trainable state and limiting GPU memory usage as the scene grows. 

\subsection{Occupancy-Guided Gaussian Initialization}\label{subsec:occ_guided_gs}

\begin{figure}[t]
    \centering
    \includegraphics[width=\linewidth]{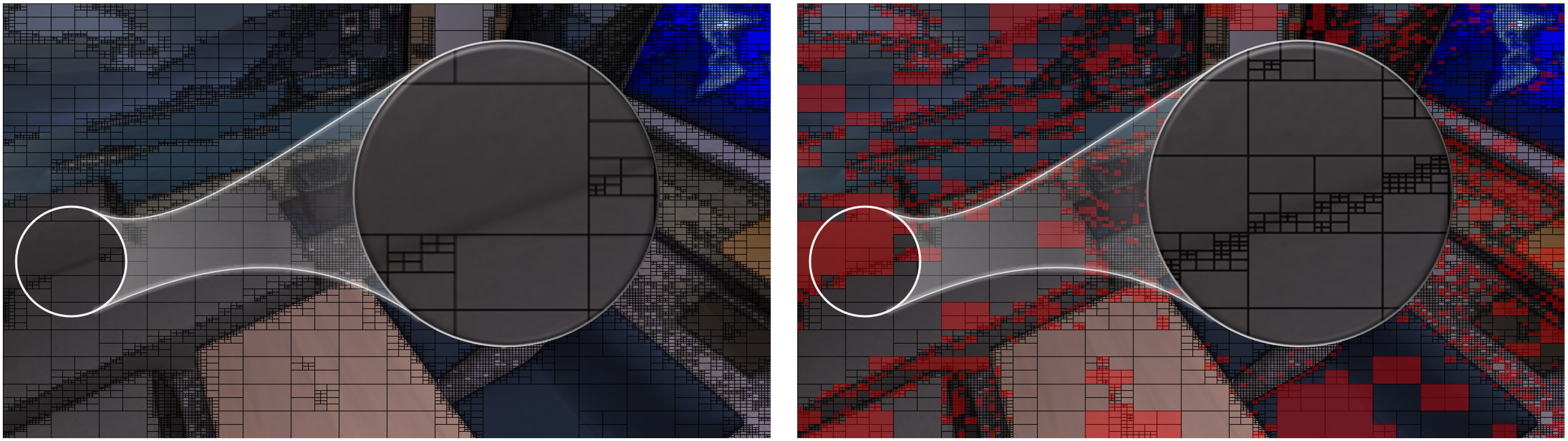}
    \caption{Comparison of quadtree-based image decomposition on \cite{straub2019replica}. Left: decomposition in \cite{wei2024gsfusion}. Right: our dual-threshold decomposition. Red regions highlight additional subdivisions that capture the image details effectively.
    }
    \label{fig:quadtree_decomposition}
\end{figure}

To improve the efficiency of 3DGS initialization, we adopt the quadtree image decomposition from \cite{wei2024gsfusion} and extend it with a dual-threshold criterion. 
A tree node is subdivided either when its interior contrast is high or when the contrast against its surrounding band is high, capturing boundaries that the interior test alone would miss.
As shown in Fig.~\ref{fig:quadtree_decomposition}, our criterion produces additional finer subdivisions around image structures that are not split by the original criterion.
Further, using summed-area tables enables $O(1)$ rectangle-sum queries, avoiding repeated pixel scans during quadtree segmentation and significantly reducing computation time.

This approach results in a set of leaf nodes $\Psi$, where each element $\psi_i = \left( \mathbf{u}_i, d_i \right) $ with pixel center $\mathbf u_i$ and depth $d_i$ proposes at most one Gaussian primitive via back-projection:
\begin{equation}
\myFrameVec{\mu}{S_m}{i}{}
=
\myFrameVec{T}{}{S_mC_t}{}
\myFrameVec{p}{C_t}{i}{}{},
\qquad
\myFrameVec{p}{C_t}{i}{}=\pi^{-1}(\myFrameVec{u}{}{i}{},d_i),
\label{eq:candidate_backproject_method}
\end{equation}
which uses the camera pose estimate $\myFrameVec{T}{}{WC_t}{}$ at time $t$ to express the current camera pose in the submap frame:
\begin{equation}
\myFrameVec{T}{}{S_mC_t}{}
=
\myFrameVec{T}{}{WS_m}{-1}
\myFrameVec{T}{}{WC_t}{}
.
\label{eq:base_render_pose_method}
\end{equation}

Out of these Gaussian primitive proposals, \gofusion accepts only candidates whose occupancy voxel in the current active local-occupancy map $\myFrameVec{O}{}{m}{}$ has integration weight one:
\begin{equation}
    \myFrameVec{G}{}{m}{\mathrm{new}} = 
    \left\{ \psi_i \in \Psi \mid d_i \text{ valid} \land w_{\myFrameVec{O}{}{m}{}} 
    \left( \myFrameVec{\mu}{S_m}{i}{} \right) = 1
    \right\},
\end{equation}
where $w_{O_m}(\cdot)$ returns the weight of the voxel corresponding to $\myFrameVec{\mu}{S_m}{i}{}$, restricting insertion to newly observed occupied voxels and avoiding redundant primitives. The accepted proposal is initialized as a flattened anisotropic 3D Gaussian to fit the surface geometry. Its local $z$-axis is aligned with the depth normal and scaled to $\exp(\myFrameScalar{s}{i,z}{})=5\times10^{-6}\,\mathrm m$, while the in-plane scales follow the quadtree footprint.


\subsection{Joint Gaussian-Pose Refinement}\label{subsec:pose_feedback}
During online mapping, \gofusion optimizes the Gaussian parameters while keeping the camera poses from \cite{boche2025okvis2x} fixed during each mapping optimization.
Joint Gaussian-pose refinement is performed at the final global 3DGS optimization stage as in \cite{wei2024gsfusion}.
All Gaussian submaps are then transformed into the world frame and merged into $\myFrameVec{G}{}{}{}$.
At refinement iteration $k$, the merged Gaussian map is rendered as
\begin{equation}
\myFrameVecHat{I}{}{t}{k}
= \mathcal{R}\!\left(
\myFrameVec{G}{}{}{},\
\Delta\myFrameVec{T}{}{t}{k}\myFrameVecBar{T}{}{C_tW}{},~
\myFrameVec{K}{}{}{}
\right),
\label{eq:render_merged_submaps_method}
\end{equation}
where $\mathcal R(\cdot)$ is the differentiable Gaussian rasterization and $\myFrameVec{K}{}{}{}$ is the camera intrinsics. The pose prior $\myFrameVecBar{T}{}{C_tW}{}=\myFrameVec{T}{}{WC_t}{-1}$ is held fixed during optimization, while the left perturbation $\Delta\myFrameVec{T}{}{t}{k}$, initialized with $\Delta\myFrameVec{T}{}{t}{0}=\myFrameVec{I}{}{4}{}$, is refined jointly with the Gaussian parameters.
We implement our CUDA rasterizer following prior work~\cite{matsuki2024gaussian}. 

The $L_1$ photometric loss
\begin{equation}
\mathcal L_{\mathrm{rgb},t}
\triangleq
\mathcal L_{\mathrm{rgb}}(\myFrameVecHat{I}{}{t}{},~\myFrameVec{I}{}{t}{})
=
\|
\myFrameVecHat{I}{}{t}{}-\myFrameVec{I}{}{t}{}
\|_1
\label{eq:photometric_loss_method}
\end{equation}
forms the basis of our mapping objective, which we can optionally extend with depth and normal supervision as
\begin{equation}
\mathcal L_{\mathrm{map},t}
=
\mathcal L_{\mathrm{rgb},t}
+\lambda_D\mathcal L_{D,t}
+\lambda_N\mathcal L_{N,t},
\label{eq:mapping_loss_method}
\end{equation}
where $\lambda_D=0.1$ and $\lambda_N=0.15$ are the empirically chosen weights.
The depth and normal terms are
\begin{equation}
\footnotesize
\mathcal{L}_{D,t} = \frac{1}{|\Omega_{D,t}|} \sum_{u} | \hat{d}_{u} - d_{u} |,
~
\mathcal{L}_{N,t} = \frac{1}{|\Omega_{N,t}|} \sum_{u} ( 1 - \hat{\alpha}_{u} \hat{\mathbf{n}}_{u}^{\top} \mathbf{n}_{u} ),
\label{eq:geometry_losses_method}
\end{equation} 
where $\Omega_{D,t}$ and $\Omega_{N,t}$ are the valid pixel sets for depth and normal supervision. $\myFrameScalar{d}{u}{}$ and $\myFrameVec{n}{}{u}{}$ denote the input depth and its derived unit normal, and $\myFrameScalarHat{d}{u}{}$, $\myFrameVecHat{n}{}{u}{}$ the rendered depth and composited normal. The accumulated pixel opacity $\myFrameScalarHat{\alpha}{u}{}$ from the current rendering is detached in the normal loss.

Following the pose optimization strategy in~\cite{matsuki2024gaussian}, we obtain the pose increment $\delta\myFrameVec{\xi}{}{t}{k} \in \mathbb{R}^6$ by backpropagating $\mathcal{L}_{\mathrm{map},t}$ through $\mathcal{R}(\cdot)$ and update the perturbation as in \cite{boche2025okvis2x}. 

The joint Gaussian-pose optimization keeps residual pose errors out of the Gaussian parameters. As a final step, we merge all submaps and refine the Gaussian parameters together with per-keyframe pose increments over the global map.


\subsection{Rasterized Gaussian-Map Fisher Approximation}\label{subsec:fisher_cache_method}
During 3DGS optimization in Sec.\,\ref{subsec:pose_feedback}, we accumulate a per-primitive image sensitivity, which measures how strongly the rendered mapping views constrain that primitive. 
Under homoscedastic Gaussian noise, this sensitivity equals the per-primitive Fisher information up to the noise scale~\cite{hanson2025pup3dgs,jiang2024fisherrf}. We use it as an uncertainty measure over the Gaussian map.

However, computing the Fisher matrix across the full primitive state $\myFrameVec{g}{}{i}{} \in \mathbb{R}^{59}$ in \eqref{eq:gaussian_param_state} is prohibitive in both memory and computation at map scale. 
We therefore use a diagonal approximation and restrict the parameter set of the $i$-th Gaussian primitive to
\begin{equation}
\myFrameVecTilde{g}{}{i}{F}
=
(
\myFrameVec{\mu}{}{i}{},~
\myFrameScalarBar{\alpha}{i}{}
)
\in \mathbb R^{4},
\label{eq:reduced_fisher_params_method}
\end{equation}
where $\myFrameScalarBar{\alpha}{i}{}=(1+\exp(-\omega_i))^{-1}$ is the per-primitive opacity obtained by applying the sigmoid function to the opacity logit $\omega_i$, as in 3DGS~\cite{kerbl2023gaussian}.

For Gaussian primitive $i$, we obtain the diagonal Fisher matrix entries by summing the diagonal Gauss-Newton contributions over RGB channel indices $c\in\{1,2,3\}$ and pixels $\myFrameVec{u}{}{}{}\in\myFrameVecHat{I}{}{v}{}$ of each rendered image view $v\in V$ as
\begin{equation}
\myFrameVec{h}{}{i}{m}
= \sum_{v,\,c,\,\myFrameVec{u}{}{}{}}
\operatorname{diag}\!\left( \myFrameVec{J}{}{}{\top}\myFrameVec{J}{}{}{} \right),
\quad
\myFrameVec{J}{}{}{} = \frac{\partial \myFrameScalarHat{I}{v,c,\myFrameVec{u}{}{}{}}{}}{\partial \myFrameVecTilde{g}{}{i}{F}}
\in \mathbb{R}^{1 \times 4},
\label{eq:ideal_fisher_method}
\end{equation}
which reduces to an elementwise square of $\myFrameVec{J}{}{}{}$. 
In \eqref{eq:ideal_fisher_method}, we use the standard front-to-back alpha compositing of Gaussian splatting, 
\begin{equation}
\small
\myFrameVecHat{I}{}{v,\myFrameVec{u}{}{}{}}{}
=
\sum_{i\in\mathcal A_v(\myFrameVec{u}{}{}{})}
\tau_{i,\myFrameVec{u}{}{}{}}\, a_{i,\myFrameVec{u}{}{}{}}\, \myFrameVec{c}{}{i,\myFrameVec{u}{}{}{}}{},
\quad
\tau_{i,\myFrameVec{u}{}{}{}}
=
\prod_{j<i}
(1 - a_{j,\myFrameVec{u}{}{}{}}),
\label{eq:method_alpha_compositing}
\end{equation}
where $\mathcal{A}_v(\myFrameVec{u}{}{}{})$ is the depth-ordered set of primitives contributing to pixel $\myFrameVec{u}{}{}{}$, $a_{i,\myFrameVec{u}{}{}{}} = \bar{\alpha}_i\Phi_{i}(\myFrameVec{u}{}{}{},\myFrameVec{\mu}{}{i}{})$ the opacity of primitive $i$ at that pixel with its projected Gaussian $\Phi_i(\cdot)$. $\tau_{i,\myFrameVec{u}{}{}{}}$ denotes the accumulated transmittance, as in \cite{kerbl2023gaussian}. 

However, evaluating \eqref{eq:ideal_fisher_method} requires the channel-wise Jacobians of every primitive contributing to a pixel. We instead reuse the rasterizer forward buffers and the existing backward pass, seeding the Fisher branch with the upstream gradient pattern $(1,1,1)$ for every rendered RGB pixel. The resulting vector-Jacobian product returns the channel-summed sensitivity of each reduced parameter directly, so that for a rendered mapping view $v$ we accumulate
\begin{equation}
\myFrameVecTilde{h}{}{v,i}{m}
= \sum_{\myFrameVec{u}{}{}{}}
\left(
\sum_c
\frac{\partial \hat{I}_{v,c,\myFrameVec{u}{}{}{}}}{\partial \myFrameVecTilde{g}{}{i}{F}}
\right)^{\!\circ 2}
\label{eq:fisher_proxy_method}
\end{equation}
with the elementwise square $(\cdot)^{\circ 2}$ . Summing the channels before squaring yields the exact diagonal Fisher of the channel-summed observation $\sum_c \hat{I}_{v,c,\myFrameVec{u}{}{}{}}$ under homoscedastic noise up to a constant scale. We therefore use $\myFrameVecTilde{h}{}{v,i}{m}$, the $i$-th row of the Fisher matrix $\myFrameVecTilde{H}{}{v}{m}\in\mathbb R_{+}^{|\mathcal G_m|\times 4}$, as a lightweight relative-sensitivity proxy of the Fisher information and evaluate it in Sec.\,~\ref{sec:fisher_eval}.

Because the Gaussian parameters change during optimization, their sensitivities accumulated from earlier views become stale. We therefore smooth the Fisher information $\myFrameVecTilde{H}{}{v}{m}$ of each rendered view $v$ over a subset of its past keyframes $v'\in V^{\mathrm{samp}}$ randomly sampled from all stored historical views $V^{\mathrm{hist}}$ of the active submap $\mathcal{M}_m$ as
\begin{equation}
\myFrameVecBar{H}{}{v}{m,\mathrm{hist}}
=
\frac{|V^{\mathrm{hist}}|}
{|V^{\mathrm{samp}}|}
\sum_{v'\in V^{\mathrm{samp}}}
\myFrameVecTilde{H}{}{v'}{m}.
\label{eq:fisher_hist_scale_method}
\end{equation}

Finally, we update the accumulated Fisher approximation $\myFrameVecHat{H}{}{m}{}$ in \eqref{eq:GOFusion_submap_state} for all primitives in $\mathcal G_m$ as
\begin{align}
\myFrameVecHat{H}{}{m}{}
\leftarrow
(1-\eta)
\left(
\myFrameVecHat{H}{}{m}{}
+
\myFrameVecTilde{H}{}{v}{m}
\right)
+
\eta\myFrameVecBar{H}{}{v}{m,\mathrm{hist}},
\label{eq:fisher_ema_method}
\end{align}
where $\eta=0.2$ is empirically selected in this work. New primitives enter with a zero row, and we initialize $\myFrameVecHat{H}{}{m}{}=\myFrameVec{0}{}{}{}$ when a Gaussian submap is created.


\subsection{Fisher-Informed Gaussian Pruning and View Selection}\label{subsec:fisher_pruning}
Using the accumulated Fisher information $\myFrameVecHat{H}{}{m}{}$ in Sec.\,\ref{subsec:fisher_cache_method}, we derive a relative uncertainty score per Gaussian primitive and show that it can be used for two different objectives: at the primitive level, the score prunes persistently unconstrained primitives, while it also identifies poorly constrained regions aggregated over space, and guides next-best-view selection (see Sec.\,\ref{subsec:experiment:nbv} for more practical details).

\subsubsection{Weakly-Constrained Gaussian Primitive Pruning}
Inspired by \cite{hanson2025pup3dgs}, we combine the position and opacity Fisher information as a coverage heuristic to prune weakly-constrained Gaussian primitives.
Using the log-scale $\myFrameVec{s}{}{i}{}$ in \eqref{eq:gaussian_param_state} and opacity $\myFrameScalarBar{\alpha}{i}{}$ in \eqref{eq:reduced_fisher_params_method}, we define
$A_i=\bigl(\frac{1}{3}\sum_{l\in\{x,y,z\}}\exp(\myFrameScalar{s}{i,l}{})\bigr)^2$,
where $\myFrameScalar{s}{i,l}{}$ denotes the log-scale along principal axis $l$ of Gaussian $i$.
The squared mean scale $A_i$ serves as an area-averaging, view-independent supporting metric, so that $\myFrameScalarBar{\alpha}{i}{}A_i$ approximates the view coverage.
The resulting importance score of each Gaussian primitive is 
\begin{equation}
\chi_i = r(\beta_i)+\lambda_{\alpha}r(\gamma_i)+r(\delta_i),
\label{eq:fisher_prune_score}
\end{equation}
with $\beta_i = \log\bigl(1 + A_i \sum_{j \in \{x,y,z\}} \hat{H}_{i,j}\bigr)$, $\gamma_i = \log(1 + \bar{\alpha}_i \hat{H}_{i,\alpha})$, and $\delta_i = \log(1 + \bar{\alpha}_i A_i)$,
where \(r(\cdot)\in[0,1]\) maps each term to its rank percentile among all primitives in the submap, placing the three terms on a common scale before weighting, and \(\lambda_{\alpha}\) weights the opacity term. The logarithm keeps the terms in a numerically stable range. Here, $\beta_i$ and $\gamma_i$ reward position and opacity information, while $\delta_i$ prevents large smooth surfaces from being over-pruned. 

For a target pruning ratio $\rho\in[0,1)$, we partition the merged Gaussian map with a voxel grid and remove the lowest-scoring fraction $\rho$ of primitives in each voxel by ascending score $\myFrameScalar{\chi}{i}{}$, keeping at least one primitive per voxel to preserve coverage in sparse regions. We evaluate this pruning strategy in Sec.\,\ref{subsec:experiment:pruning}.

\subsubsection{Next-Best-View Selection}
To select the next-best-view for robot navigation, we use Fisher information to favor views whose rendered images are most sensitive to the least-constrained Gaussian parameters.
Let $\mathcal M^{\mathrm{query}}\subseteq\mathcal M$ denote the associated submaps used to score the query view $v$, and let $\myFrameScalarHat{H}{i,j}{}$ be the entry of $\myFrameVecHat{H}{}{m}{}$ for primitive $i$ and component $j\in\{x,y,z,\alpha\}$, with the submap index omitted for readability. We use the regularized inverse $U_{i,j}
=
\frac{1}{\myFrameScalarHat{H}{i,j}{}+\epsilon}$ as a relative uncertainty score, where $\epsilon=10^{-6}$ caps the score at $1/\epsilon$ for near-zero entries. Larger $U_{i,j}$ indicates weaker accumulated image constraints.

Consequently, we use this relative uncertainty score $U_{i,j}$ to weight the single-view Fisher information $\myFrameVecTilde{h}{}{v,i}{m}$ in \eqref{eq:fisher_proxy_method} across all Gaussian primitives in $\mathcal G_m$ for a candidate view $v$. 
Following \cite{jiang2024fisherrf}, we scale each component of $\myFrameVecTilde{h}{}{v,i}{m}$ by its corresponding inverse 
historical information to obtain the per-primitive relative information-gain score $\Gamma_i(v)$:
\begin{equation}
\Gamma_i(v)
= \frac{1}{3}\sum_{j\in\{x,y,z\}}
U_{i,j}\bigl[\myFrameVecTilde{h}{}{v,i}{m}\bigr]_j
+ U_{i,\alpha}\bigl[\myFrameVecTilde{h}{}{v,i}{m}\bigr]_{\alpha},
\label{eq:fisher_view_gain_method}
\end{equation}
 where $[\cdot]_j$ selects the $j$-th component of the corresponding parameter. 

Across the queried submaps $\mathcal M_m\in\mathcal M^{\mathrm{query}}$, we compute $\Gamma(v)=\operatorname{mean}_{m,i:\,\Gamma_i(v)\ne0}\Gamma_i(v)$, then normalize it as $\Gamma(v)\leftarrow\Gamma(v)/(1+\Gamma(v))$ to obtain the final next-best-view score.
In Sec.\,\ref{subsec:experiment:nbv}, we demonstrate how this score guides view selection during autonomous drone exploration to refine the Gaussian map. We adopt the submap-based drone exploration framework~\cite{papatheodorou2025efficient}, and augment its candidate-view utility by adding the normalized Fisher score $\Gamma(v)$ as a weighted term alongside its existing occupancy information gain. 
The planner then favors view candidates with a higher $\Gamma(v)$ and emits a collision-free path to the robot.
\newcommand{\maketabletumrgbd}{
\begin{table}[t]
\centering
\caption{Benchmark of rendering quality and map compactness on
TUM-RGBD~\cite{sturm2012benchmark} with ground-truth poses, averaged over 10
runs. All methods re-evaluated on full images, not cited from the original
papers. (+D/N) denotes explicit depth and normal supervision in the 3DGS
objective. \textbf{Best} and \underline{second-best} in bold and underlined.}
\label{tab:gt_pose_mapping}
{\scriptsize
\setlength{\tabcolsep}{2.5pt}
\begin{tabular*}{\columnwidth}{@{\extracolsep{\fill}}l cccccc@{}}
\toprule
Method & PSNR\,$\uparrow$ & SSIM\,$\uparrow$ & LPIPS\,$\downarrow$ & \makecell{Mapp.\\FPS\,$\uparrow$} & \makecell{\#Gauss.\\($10^6$)\,$\downarrow$} & \makecell{VRAM\\(GB)\,$\downarrow$} \\
\midrule
GSFusion~\cite{wei2024gsfusion}      & 20.23 & 0.708 & 0.357 & 18.17 & \textbf{0.10} & 2.1 \\
RTG-SLAM~\cite{peng2024rtgslam} & 18.47 & 0.708 & 0.470 & 10.64 & 0.31 & 2.4 \\
GPS-SLAM~\cite{peng2025gpsslam}      & 18.78 & 0.652 & 0.393 & \textbf{103.7} & 0.51 & 8.8 \\
PhotoSLAM~\cite{huang2024photoslam}  & 21.30 & 0.748 & 0.205 & \underline{34.57} & \underline{0.11} & \textbf{1.2} \\
GS-ICP SLAM~\cite{ha2024rgbdgsicp}   & 18.92 & 0.688 & 0.304 & 30.02 & 1.47 & 2.2 \\
SplaTAM~\cite{keetha2024splatam}& 19.45 & \textbf{0.801} & 0.336 & 2.40 & 1.91 & 3.9 \\
\midrule
Ours        & \textbf{21.84} & \underline{0.754} & \textbf{0.192} & 32.86 & 0.53 & \underline{1.9} \\
Ours (+D/N) & \underline{21.55} & 0.748 & \underline{0.194} & 31.35 & 0.52 & 3.2 \\
\bottomrule
\end{tabular*}
}
\end{table}
}

\newcommand{\maketableuhumans}{
\begin{table}[!t]
\caption{Benchmark of rendering quality and SLAM performance on uHumans2~\cite{rosinol2021kimera}, averaged over 10 runs (median for ATE). (+D/N) denotes explicit depth/normal supervision in the 3DGS optimization objective. \textbf{Best} and \underline{second-best} results are bold and underlined. OOM: out-of-memory.}
\label{tab:uhumans_cross_method}
\centering
{\scriptsize
\setlength{\tabcolsep}{2.5pt}
\renewcommand{\arraystretch}{0.9}
\begin{tabular*}{\columnwidth}{@{\extracolsep{\fill}}llcccccccc@{}}
\toprule
Seq. & Metric
& \rotatebox{90}{\makecell[l]{RTG-\\SLAM~\cite{peng2024rtgslam}}}
& \rotatebox{90}{\makecell[l]{GPS-\\SLAM~\cite{peng2025gpsslam}}}
& \rotatebox{90}{\makecell[l]{Photo-\\SLAM~\cite{huang2024photoslam}}}
& \rotatebox{90}{\makecell[l]{GS-ICP \\SLAM~\cite{ha2024rgbdgsicp}}}
& \rotatebox{90}{\makecell[l]{LoopSplat\\~\cite{zhu2025loopsplat}}}
& \rotatebox{90}{\makecell[l]{DiskChun\\-GS~\cite{feldmann2026diskchungs}}}
& \rotatebox{90}{\makecell[l]{\best{Ours}}}
& \rotatebox{90}{\makecell[l]{\best{Ours}\\(+D/N)}}\\
\midrule
\multirow{7}{*}{\makecell[l]{Apart.\\49.4\,m}}
& PSNR\,$\uparrow$
& 26.04 & 14.79 & 20.27 & 23.67 & 25.24 & 29.15
& \best{33.09} & \second{31.96} \\
& SSIM\,$\uparrow$
& 0.894 & 0.284 & 0.730 & 0.816 & 0.828 & 0.866
& \best{0.942} & \second{0.928} \\
& LPIPS\,$\downarrow$
& 0.330 & 0.383 & 0.392 & 0.335 & 0.324 & 0.110
& \best{0.062} & \second{0.076} \\
& Full-FPS\,$\uparrow$
& 9.66 & \second{42.35} & \best{57.13} & 10.62 & 0.59 & 10.50
& 35.23 & 30.54 \\
& ATE\,$\downarrow$
& \best{0.007} & 0.140 & 0.139 & 1.581 & 0.046 & 0.016
& 0.015 & \second{0.014} \\
& \#Gauss.\,($10^6$)\,$\downarrow$
& \best{0.14} & 0.35 & 0.31 & 2.61 & 4.50 & \second{0.18}
& 0.69 & 0.69 \\
& VRAM\,(GB)\,$\downarrow$
& 4.5 & 5.4 & 5.2 & 11.9 & 42.8 & 4.9
& \best{2.4} & \second{4.2} \\
\midrule
\multirow{7}{*}{\makecell[l]{Office\\264\,m}}
& PSNR\,$\uparrow$
& 22.64
& \multirow{7}{*}{\rotatebox{90}{GPU OOM}}
& 14.40
& \multirow{7}{*}{\rotatebox{90}{GPU OOM}}
& \multirow{7}{*}{\rotatebox{90}{CPU OOM}}
& 24.49
& \best{31.14} & \second{29.26} \\
& SSIM\,$\uparrow$
& 0.813 & & 0.589 & & & 0.794
& \best{0.939} & \second{0.910} \\
& LPIPS\,$\downarrow$
& 0.421 & & 0.631 & & & 0.263
& \best{0.053} & \second{0.076} \\
& Full-FPS\,$\uparrow$
& 8.30 & & \best{46.89} & & & 12.90
& \second{24.60} & 21.51 \\
& ATE\,$\downarrow$
& 0.219 & & 0.236 & & & 3.198
& \second{0.062} & \best{0.052} \\
& \#Gauss.\,($10^6$)\,$\downarrow$
& \second{0.42} & & 1.92 & & & \best{0.39}
& 2.31 & 2.32 \\
& VRAM\,(GB)\,$\downarrow$
& 17.5 & & 40.7 & & & \second{8.3}
& \best{6.3} & 10.8 \\
\midrule
\multirow{7}{*}{\makecell[l]{Subway\\508.7\,m}}
& PSNR\,$\uparrow$
& 23.11
& \multirow{7}{*}{\rotatebox{90}{GPU OOM}}
& \multirow{7}{*}{\rotatebox{90}{diverged}}
& \multirow{7}{*}{\rotatebox{90}{GPU OOM}}
& \multirow{7}{*}{\rotatebox{90}{CPU OOM}}
& 26.11
& \best{29.99} & \second{28.27} \\
& SSIM\,$\uparrow$
& 0.782 & & & & & 0.780
& \best{0.902} & \second{0.859} \\
& LPIPS\,$\downarrow$
& 0.479 & & & & & 0.223
& \best{0.085} & \second{0.125} \\
& Full-FPS\,$\uparrow$
& 7.76 & & & & & \best{12.95}
& \second{11.95} & 11.11 \\
& ATE\,$\downarrow$
& 0.504 & & & & & 26.717
& \second{0.180} & \best{0.180} \\
& \#Gauss.\,($10^6$)\,$\downarrow$
& \best{0.56} & & & & & \second{1.35}
& 3.90 & 3.90 \\
& VRAM\,(GB)\,$\downarrow$
& 25.8 & & & & & 22.8
& \best{9.1} & \second{13.5} \\
\midrule
\multirow{7}{*}{\makecell[l]{Neighb.\\337.8\,m}}
& PSNR\,$\uparrow$
& 19.75
& \multirow{7}{*}{\rotatebox{90}{GPU OOM}}
& \multirow{7}{*}{\rotatebox{90}{diverged}}
& \multirow{7}{*}{\rotatebox{90}{GPU OOM}}
& \multirow{7}{*}{\rotatebox{90}{CPU OOM}}
& 20.14
& \best{25.46} & \second{24.35} \\
& SSIM\,$\uparrow$
& 0.710 & & & & & 0.661
& \best{0.829} & \second{0.781} \\
& LPIPS\,$\downarrow$
& 0.533 & & & & & 0.387
& \best{0.163} & \second{0.209} \\
& Full-FPS\,$\uparrow$
& \second{7.57} & & & & & \best{12.72}
& 7.12 & 6.74 \\
& ATE\,$\downarrow$
& 8.721 & & & & & 5.831
& \second{0.279} & \best{0.266} \\
& \#Gauss.\,($10^6$)\,$\downarrow$
& \best{0.35} & & & & & \second{0.50}
& 1.99 & 2.01 \\
& VRAM\,(GB)\,$\downarrow$
& 16.0 & & & & & 8.7
& \best{4.9} & \second{7.4} \\
\bottomrule
\end{tabular*}}
\end{table}
}
\newcommand{\maketableflight}{
\begin{table}[!t]
\centering
\caption{Benchmark of rendering quality and map compactness on the real-world
flight sequences using estimated poses, averaged over 16 sequences. (+D/N)
denotes explicit depth and normal supervision in the 3DGS objective. \textbf{Best} and \underline{second-best} in
bold and underlined.}
\label{tab:realworld_16_run}
{\scriptsize
\setlength{\tabcolsep}{2.5pt}
\begin{threeparttable}
\begin{tabular*}{\columnwidth}{@{\extracolsep{\fill}}c|l ccccc@{}}
\toprule
\multicolumn{1}{c}{} & Method & PSNR\,$\uparrow$ & SSIM\,$\uparrow$ & LPIPS\,$\downarrow$
& \makecell{\#Gauss.\\($10^6$)\,$\downarrow$} 
& \makecell{VRAM\\(GB)\,$\downarrow$} \\
\midrule
\multirow{9}{*}{\rotatebox{90}{Offline}}
 & GSFusion~\cite{wei2024gsfusion}
    & 24.23 & 0.701 & 0.443 & 0.40 & 7.7 \\
 & RTG-SLAM~\cite{peng2024rtgslam}
    & 15.85 & 0.478 & 0.693 & 0.46 & 8.6 \\
 & GPS-SLAM~\cite{peng2025gpsslam}
    & 7.19 & 0.125 & 0.970 & \best{0.31} & 14.3 \\
 & PhotoSLAM~\cite{huang2024photoslam}
    & 23.35 & 0.642 & 0.524 & \second{0.36} & \best{6.1} \\
 & GS-ICP SLAM~\cite{ha2024rgbdgsicp}
    & 22.36 & 0.758 & 0.267 & 2.92 & 17.5 \\
 & LoopSplat~\cite{zhu2025loopsplat}
    & \multicolumn{5}{c}{CPU OOM} \\
 & DiskChunGS~\cite{feldmann2026diskchungs}$^{*}$
    & 23.48 & 0.679 & 0.478 & 1.14 & 17.8 \\
 & Ours
    & \best{26.35} & \best{0.808} & \second{0.167} & 3.88 & 9.0 \\
 & Ours (+D/N)
    & \second{26.04} & \second{0.801} & 0.186 & 3.89 & 10.4 \\
\midrule
\multirow{3}{*}{\rotatebox{90}{Online~}}

 & Ours$^{1\dagger}$ (pre-global) 
    & 16.35 & 0.486 & 0.611 & 2.44 & \second{7.1} \\
 & Ours$^{2\dagger}$ (post-global)
    & 24.86 & 0.740 & 0.259 & 2.44 & \second{7.1} \\
 & Ours$^{3\dagger}$ (post-global)
    & 25.03 & 0.782 & \best{0.157} & 2.44 & \second{7.1} \\
\bottomrule
\end{tabular*}
\begin{tablenotes}[flushleft]
\fontsize{7}{8.4}\selectfont
\item[$^{*}$] DiskChunGS~\cite{feldmann2026diskchungs} only completed 12 of 16 runs due to GPU OOM.
\item[$^{\dagger}$] Live results with depth and normal supervision. 
\item[1] With live SLAM poses and real-time, incremental 3DGS refinement (pre-global).
\item[2] With live SLAM poses and global 3DGS refinement (post-global).
\item[3] With BA-refined poses and global 3DGS refinement (post-global).
\end{tablenotes}
\end{threeparttable}
}
\end{table}

}
\newcommand{\maketableablation}
{
\begin{table}[!t]
\caption{Ablation of map format and joint Gaussian-pose optimization on uHumans2~\cite{rosinol2021kimera}, averaged over 10 independent runs across all sequences. \textbf{Best} and \underline{second-best} results are bold and underlined.}
\label{tab:ablation_submap_gradient}

\centering
{\scriptsize
\setlength{\tabcolsep}{4pt}
\renewcommand{\arraystretch}{1.0}
\begin{tabular*}{\columnwidth}{@{\extracolsep{\fill}}lcccc@{}}
\toprule
\multirow{2}{*}{Metric}
& \multicolumn{2}{c}{Single Map}
& \multicolumn{2}{c}{Multiple Submaps} \\
\cmidrule(lr){2-3}\cmidrule(lr){4-5}
& w/o. Joint Opt. & w. Joint Opt. & w/o. Joint Opt. & w. Joint Opt. \\
\midrule
\ac{psnr}\,$\uparrow$
& 25.01 & 25.06 & \second{29.68} & \best{29.92} \\
\ac{ssim}\,$\uparrow$
& 0.774 & 0.775 & \second{0.898} & \best{0.903} \\
\ac{lpips}\,$\downarrow$
& 0.274 & 0.269 & \second{0.096} & \best{0.091} \\
\ac{ate}\,$\downarrow$
& 0.150 & \best{0.124} & \second{0.134} & 0.134 \\
Full-FPS\,$\uparrow$
& 17.95 & 17.97 & \second{19.60} & \best{19.72} \\
\#Gauss.\,($10^6$)\,$\downarrow$
& \best{0.60} & \second{0.61} & 2.23 & 2.22 \\
VRAM\,(GB)\,$\downarrow$
& 15.7 & 15.7 & \second{6.6} & \best{5.7} \\
\bottomrule
\end{tabular*}}
\end{table}
}
\section{Experiment and Results}\label{sec:experiments}

\subsection{Experimental Setup}
We evaluate the proposed method on two public datasets and a real-world flight, comparing against state-of-the-art baselines throughout. 
TUM RGB-D~\cite{sturm2012benchmark} provides real-world indoor sequences with motion-capture ground-truth trajectories, while the uHumans2~\cite{rosinol2021kimera} offers larger-scale synthetic environments with dense ground-truth geometry. The flight experiments additionally demonstrate real-time exploration of unknown, larger-scale environments, with the proposed Fisher information guiding view selection to refine the reconstruction.

All experiments run on a workstation with an Intel Core Ultra 9-285K CPU, 96 GB of RAM, and an NVIDIA RTX 4090 GPU with 48 GB of VRAM, which also serves as the ground station for the outdoor flight\footnote{A fully onboard configuration at reduced resolution is shown in the supplementary video due to space limitations.}. In the real-world experiment, we used a self-built quadcopter with a MicroStrain 3DM-GX5-25 IMU, a custom stereo rig of two ZED X One GS cameras, and a Jetson Orin NX with 16 GB of unified memory, running visual-inertial SLAM alongside a PX4 flight controller.

In contrast to the public datasets, which provide low-resolution RGB and depth images at $640\times480$, our outdoor-flight experiments use $1920\times1200$ images\footnote{Depth images are processed by the ZED SDK (v5.4, \texttt{NEURAL\_PLUS} mode). See \href{https://www.stereolabs.com/en-de/developers}{stereolabs.com}.} at \SI{15}{Hz} together with IMU measurements at \SI{200}{Hz}. For DiskChunGS~\cite{feldmann2026diskchungs}, we supply depth directly to bypass its stereo-depth network without loading or executing it.

Unlike several baselines that evaluate only on valid-depth regions, we compute \ac{psnr}, \ac{ssim} and \ac{lpips} over the full image for all methods, since regions without depth returns are still rendered. This is why our reproduced scores for some baselines fall below the numbers reported in their papers. We further measure \ac{ate} RMSE, the number of Gaussian primitives, and peak GPU VRAM usage. Mapping \ac{fps} (Mapp. FPS) reports mapping throughput and full-system \ac{fps} (Full-FPS) the full tracking-and-mapping pipeline, both excluding the post-\ac{slam} refinement.


\subsection{Evaluation of Mapping Quality}

\subsubsection{TUM RGB-D}\label{res:tum_rgbd}
On TUM RGB-D~\cite{sturm2012benchmark}, we first evaluate reconstruction quality using ground-truth camera poses to isolate Gaussian mapping from trajectory estimation. Numbers are averaged over ten runs across all scenes in Table~\ref{tab:gt_pose_mapping}. Both \gofusion configurations, with and without depth and normal supervision, take the top two places on PSNR and LPIPS at competitive mapping FPS. Although our image decomposition produces more primitives than \cite{wei2024gsfusion}, the memory footprint grows far more slowly, with 5.3 times as many primitives at 0.9 times the VRAM. The reason our SSIM falls behind \cite{keetha2024splatam} is likely its D-SSIM loss term and a three-times-larger primitive count. Interestingly, depth and normal supervision trade a small drop in RGB scores for better geometric consistency and map completeness. We further discuss this observation in Sec.\,\ref{subsec:ablation_depth_normal}.

\subsubsection{uHumans2}
In extension to Sec.\,\ref{res:tum_rgbd}, we next evaluate our method against a comparable set of baselines on uHumans2~\cite{rosinol2021kimera} as a full-mapping system with estimated camera poses, in which SLAM also affects reconstruction quality. As shown in Table~\ref{tab:uhumans_cross_method} and Fig.\,\ref{fig:uhumans2_demo}, our method leads on all rendering metrics across all four sequences and achieves the lowest ATE on the three long ones. Although we generate more Gaussian primitives, our memory footprint remains lower than that of \cite{peng2024rtgslam} and \cite{feldmann2026diskchungs}. 
This highlights the benefit of our system design that anchors occupancy and Gaussian primitives to share submap frames. Confining the optimization to the active submap guarantees \gofusion's scalability and enables consistent reconstruction across large environments where several baselines fail to complete sequences or suffer from trajectory drift. As observed in Sec.\,\ref{res:tum_rgbd}, adding depth and normal supervision during 3DGS optimization does not necessarily improve rendering quality, since it regularizes the reconstruction toward geometric consistency.

\maketabletumrgbd
\maketableuhumans
\maketableflight

\subsubsection{Self-Collected Outdoor Flying Data}
We benchmark all methods on 16 self-collected outdoor flight sequences\footnote{The supplementary video shows the outdoor flight experiment.} with challenging real-world illumination. Table~\ref{tab:realworld_16_run} reports the mapping performance in two settings. In the offline setting, all methods use camera poses from our SLAM output after bundle adjustment (BA)\footnote{For a fair comparison of rendering quality, we supply our BA-refined poses to all baselines in offline evaluation in Table~\ref{tab:realworld_16_run}, since their own tracking produces large trajectory errors or diverges entirely in Table~\ref{tab:uhumans_cross_method}.}, which are held fixed as pseudo-GT poses and also fixed during global 3DGS refinement. In the online setting, we report metrics obtained with live SLAM pose estimates and incremental 3DGS optimization before and after global 3DGS refinement. SLAM runs on the resource-limited onboard platform, while the 3DGS optimization was executed on a ground station.

As in the previous sections, a similar conclusion on mapping performance can be drawn from the offline evaluation, with \cite{zhu2025loopsplat} and \cite{feldmann2026diskchungs} additionally failing to complete all sequences. Our offline configuration with RGB-only supervision performs best on PSNR and LPIPS, as shown in Sec.\,\ref{res:tum_rgbd}. 

For online, real-time drone navigation, our method adaptively skips frames to avoid blocking, thereby consuming fewer frames than offline baselines. Despite this, it already surpasses most offline baselines in rendering quality with global 3DGS refinement across all submaps (i.e.\,post-global), whereas those baselines were evaluated offline with BA-refined poses. Under identical online conditions, it also surpasses most offline baselines in memory usage with BA-refined poses. Since our incremental 3DGS optimization (i.e.\,pre-global) runs only on the current submap to bound computation and keep the system real-time, the rendering quality drops but still exceeds~\cite{peng2025gpsslam, peng2024rtgslam}. This is sufficient during flight, where navigation plans on the occupancy layer and view selection relies on Fisher information rather than on rendering fidelity (see Sec.\,\ref{sec:fisher_eval}). This quality gap closes after a few passes of global optimization.
 
In addition, we found that a larger Gaussian count, which results from denser keyframe selection and a finer occupancy resolution (e.g., \SI{0.1}{m}), does not translate into a proportional memory cost, since offline \gofusion holds roughly ten times as many primitives as \cite{huang2024photoslam} at less than twice its VRAM. 


\subsection{Ablation Studies}\label{subsec:ablation}

\subsubsection{Submapping and Joint Optimization}\label{subsec:ablation_submap_joint_opt}
We study the submap-based map representation and joint Gaussian-pose optimization on uHumans2~\cite{rosinol2021kimera}, using only RGB supervision. As shown in Table~\ref{tab:ablation_submap_gradient}, partitioning the map into submaps improves every rendering metric, slightly raises full-system FPS, and reduces peak memory, since only the active submap carries training buffers. Because each submap covers a small region, its local Gaussians stay
consistent with the underlying geometry, where a global pose correction is applied to all of them at once. Besides, the joint optimization yields further gains across all rendering metrics at no cost in full-system FPS, and lowers the ATE of the single map but not of the submaps, whose poses are already well constrained by loop closure and submap alignment.

\maketableablation

\subsubsection{Depth and Normal Supervision for 3DGS}\label{subsec:ablation_depth_normal}
\begin{figure}[!t]
\centering
\includegraphics[width=0.98\linewidth]{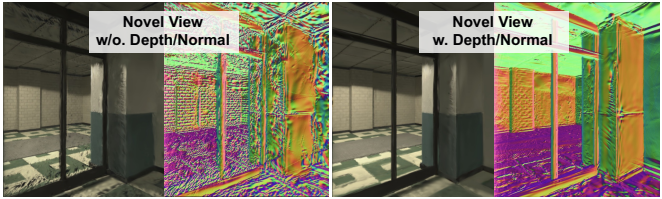}
\caption{
Qualitative novel-view comparison of 3DGS optimization without (left) and with (right) depth and normal supervision on uHumans2~\cite{rosinol2021kimera}.
}
\label{fig:abl_depth_normal}
\end{figure}
We further investigate the impact of geometric supervision through depth and normals in \eqref{eq:mapping_loss_method}. Fig.\,\ref{fig:abl_depth_normal} illustrates exemplary RGB renderings alongside their rendered normals from novel viewpoints on uHumans2~\cite{rosinol2021kimera}\footnote{Since uHumans2~\cite{rosinol2021kimera} provides only training views, we cannot report quantitative novel view synthesis results for this ablation study. Instead, we present an extensive qualitative comparison across all benchmark methods with and without depth/normal supervision in the supplementary video.}. Accordingly, depth and normal supervision consistently maintain surface geometry and substantially improve novel viewpoints, while the RGB renderings remain comparable. As discussed in \cite{park2025dropgaussian}, regularizing the 3DGS optimization toward geometric consistency prevents overfitting to the training-view appearance, which matters more for robot navigation than training-view rendering quality. However, this comes at the cost of a slight drop in rendering quality on the training views in our experiments, as shown in Table \ref{tab:uhumans_cross_method}. Consequently, the weights for depth and normal loss terms in \eqref{eq:mapping_loss_method} have to be carefully balanced.



\subsection{Evaluation of Gaussian-Map Fisher Approximation}
\label{sec:fisher_eval}

\subsubsection{Pruning of Weakly-Constrained Gaussians}
\label{subsec:experiment:pruning}
In this experiment, we evaluate the proposed online Gaussian-map Fisher approximation as a pruning heuristic to remove weakly constrained Gaussians (see Sec.\,\ref{subsec:fisher_pruning}), comparing rendering quality before and after pruning on four uHumans2 sequences~\cite{rosinol2021kimera}. We evaluate our pruning heuristics against two baseline pruning scores: a random pruning strategy that replaces the entries of $\myFrameVecHat{H}{}{m}{}$ in \eqref{eq:fisher_ema_method} with random values from a fixed seed, and an offline score recomputed over all keyframes after merging the submaps as in \cite{hanson2025pup3dgs}, which serves as the reference pruning strategy.  

\begin{figure}[!t]
\centering
\includegraphics[width=0.98\linewidth]{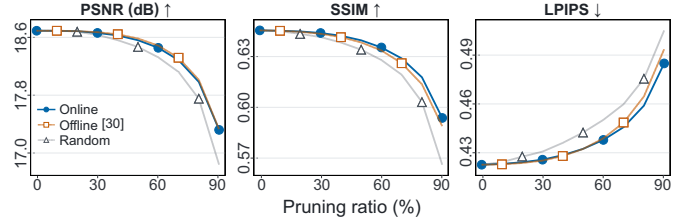}
\caption{
Comparison of three Gaussian pruning strategies evaluated on the live pre-global map with online SLAM poses on uHumans2~\cite{rosinol2021kimera}.
}
\label{fig:fisher_pruning}
\end{figure}

Fig.\,\ref{fig:fisher_pruning} shows rendering quality as a function of the pruning ratio during online Gaussian mapping. Our online Gaussian pruning using the proposed Fisher approximation achieves pruning quality comparable to the offline reference, while both clearly outperform the random baseline as pruning becomes more aggressive. Since the offline reference is only available after the session ends, our online Fisher-based heuristic offers a practical alternative for Gaussian pruning.

\subsubsection{Next-Best-View Selection}
\label{subsec:experiment:nbv}
\begin{figure}[!t]
    \centering
    \includegraphics[width=\linewidth]{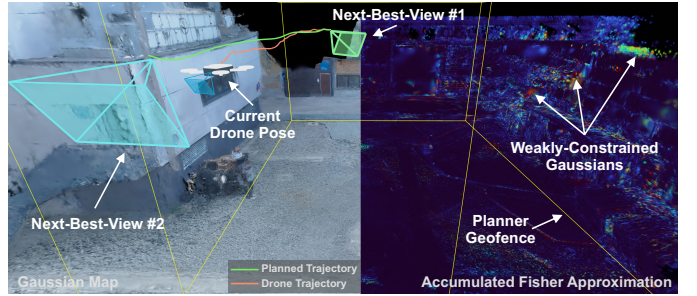}
    \caption{Real-world autonomous drone flight with two next-best-view candidates. Camera frustums (light blue and green) indicate the planned next-best-views, directed toward regions with low Fisher entries (warm color). The yellow frame marks the planner's geofence, keeping the drone out of cluttered areas during autonomous flights.
    }
    \label{fig:realworld_nbv_scene}
\end{figure}

We then show that the Gaussian-map Fisher approximation can also select the next-best-view for map refinement in a real-world autonomous flight, while the occupancy map supports collision-free path planning. To achieve the best mapping performance, the visual-inertial \ac{slam} runs onboard, while the ground station maintains the hybrid map and evaluates candidate views offboard. Fig.\,\ref{fig:realworld_nbv_scene} provides a scene overview while the drone was transiting between two next-best-view candidates. 

To examine how a selected view improves reconstruction, we compare maps reconstructed from observations accumulated before and after its visit, using the same fixed final estimated trajectory.
Both maps are rendered from the same candidate camera pose and intrinsics.

\begin{figure}[!t]
\centering
\includegraphics[width=\linewidth]{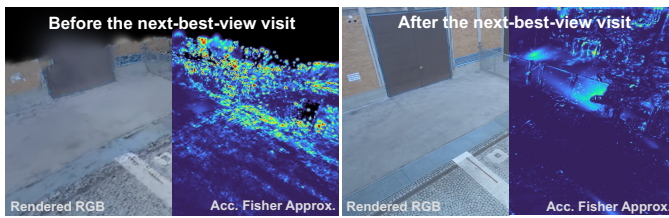}
\caption{
Map before (left) and after (right) visiting the next-best-view during the flight. The rendered RGBs are from the same candidate pose. Warmer colors in the Fisher map indicate higher relative uncertainty.
}
\label{fig:realworld_nbv_refinement}
\end{figure}

As shown in \figref{fig:realworld_nbv_refinement}, the visit improved both local detail and reconstruction completeness.
The displayed high-response regions in the Fisher uncertainty map also become more localized, with remaining responses concentrated around vegetation and object boundaries.
These qualitative results support the ability of Fisher-informed next-best-view selection to identify informative views that improve 3DGS reconstruction quality and reduce map uncertainty.

\section{Conclusion}\label{sec:conclusion}
We present \gofusion, a hybrid mapping method that anchors 3D Gaussians to local-occupancy submaps from visual-inertial SLAM to support real-time robot navigation. Beyond spatial awareness for path planning and high-fidelity appearance, it enables Fisher-guided Gaussian pruning and next-best-view selection. Across our extensive evaluations and a real-world autonomous flight, \gofusion achieves higher rendering quality than existing Gaussian \ac{slam} systems with bounded GPU memory usage and successfully completes sequences that several baselines fail to complete.

However, our method still assumes accurate and robust depth estimation, requiring careful tuning of the depth and normal supervision during 3DGS optimization. To maintain the computational efficiency on resource-constrained devices, optimizing one submap at a time limits the online rendering quality where neighboring submaps overlap or are misaligned, which would become more pronounced in semi-static or dynamic environments. 
Future work will incorporate depth uncertainty into the mapping objective, extend the optimization across neighboring submaps, and use semantic cues to handle dynamic environments.

\section*{Acknowledgments}
OpenAI Codex and ChatGPT assisted with experimental code development, experiment execution, and language polishing. The authors take full responsibility. 

This work was funded by the Technical University of Munich (MIRMI Seed Fund), supported by Huawei Technologies. The authors thank Dr. Massimiliano Maule, Hongjia Su, Dr. Xueliang Shi, Joseph Eichinger, Dr. Xueli An, and Lei Lu from Huawei Technologies for their support.



\bibliographystyle{IEEEtran}
\bibliography{reference}
\end{document}